\documentclass[letterpaper, 10 pt, conference,final]{ieeeconf}  

\IEEEoverridecommandlockouts                              

\usepackage{bm}
\usepackage{tikz}
\usepackage{algorithm}
\usepackage{algpseudocode}
\definecolor{black}{rgb}		{0.0, 0.0, 0.0}
\definecolor{white}{rgb}		{1.0, 1.0, 1.0}
\definecolor{yellow}{rgb}		{1.0, 1.0, 0.8}
\definecolor{red}{rgb}			{0.6, 0.0, 0.2}
\definecolor{blue}{rgb}		{0.0, 0.2, 0.5}
\definecolor{green}{rgb}		{0.6, 0.8, 0.8}
\definecolor{dark_green}{RGB} {0, 140, 0}
\definecolor{gold}{rgb}		{0.6, 0.4, 0.1}
\definecolor{grey}{RGB}{0,0,0}
\definecolor{Gray}{gray}{0.8}
\definecolor{MediumGray}{gray}{0.9}
\definecolor{LightGray}{gray}{0.98}
\definecolor{LightCyan}{rgb}{0.88,1,1}
\definecolor{purple}{RGB}{128,0,128}
\definecolor{sl_blue}{RGB}{47, 60, 105}
\definecolor{orange}{RGB}{255,165,0}
\definecolor{Gray}{gray}{0.85}

\usepackage{amssymb}
\usepackage{amsmath}
\let\labelindent\relax
\usepackage{enumitem}
\usepackage{makecell}
\usepackage{changes}
\definechangesauthor[name={Greg}, color=blue]{GP}
\definechangesauthor[name={Amir}, color=purple]{AA}
\definechangesauthor[name={Daniel}, color=red]{DK}

\newcommand{\eg}{e.\,g.,\ }

\usepackage{amsmath,amssymb,amsfonts}
\usepackage{breqn}
\DeclareMathOperator{\E}{\mathbb{E}}

\usepackage[english]{babel}
\usepackage[bookmarks=false]{hyperref}
\addto\extrasenglish{%
}

\addto\extrasenglish{ 	 	 	 }

\addto\extrasenglish{%
}

\usepackage{booktabs}
\usepackage{adjustbox}
\usepackage{multirow}
\usepackage{tabularx}

\usepackage{mathtools}

\DeclarePairedDelimiterX{\infdivx}[2]{(}{)}{%
  #1\;\delimsize\|\;#2%
}
\newcommand{\infdiv}{D_{KL}\infdivx}

\usepackage[figuresright]{rotating}
\usepackage{pdflscape}
\usepackage{float}
\usepackage{cleveref}

\usepackage{hyperref}
\hypersetup{
    colorlinks=true,
    linkcolor=black,
    citecolor=blue,
    filecolor=black,
    urlcolor=black
}

\usepackage{todonotes}
\setuptodonotes{inline}

\title{\LARGE \bf Data Valuation for Offline Reinforcement Learning}

\author{Amir Abolfazli, Gregory Palmer and Daniel Kudenko
\thanks{L3S Research Center, Leibniz University Hannover, Germany  {\tt\small \{abolfazli, gpalmer, kudenko\}@l3s.de}}%
}

\begin{document}

\maketitle
\thispagestyle{empty}
\pagestyle{empty}

\begin{abstract}
The success of deep reinforcement learning (DRL)
hinges on the availability of training data, which is typically
obtained via a large number of environment interactions. In many real-world scenarios, costs and risks are associated with gathering these data. The field of offline reinforcement learning addresses these issues through outsourcing the collection of
data to a domain expert or a carefully monitored program
and subsequently searching for a batch-constrained optimal policy. With the emergence of data markets, an alternative to constructing a dataset in-house is to purchase external data. However, while state-of-the-art offline reinforcement learning approaches have shown a lot of promise, they currently rely on carefully constructed datasets that are well aligned with
the intended target domains. This raises questions regarding the transferability and robustness of an offline reinforcement learning agent trained on externally acquired data. In this paper, we empirically evaluate the ability of the current state-of-the-art offline reinforcement learning approaches to coping with the source-target domain mismatch within two MuJoCo environments, finding that current state-of-the-art offline reinforcement learning algorithms underperform in the target domain. To address this, we propose data valuation for offline reinforcement learning (DVORL), which allows us to identify relevant
and high-quality transitions, improving the performance and transferability of policies learned by offline reinforcement learning algorithms. The
results show that our method outperforms offline reinforcement learning baselines
on two MuJoCo environments.
\end{abstract}

\section{Introduction}
Offline Reinforcement Learning (RL) -- also known as batch-constrained RL -- is a class of RL methods that requires the agent to learn from a static dataset of pre-collected experiences without further environment interaction~\cite{lange2012batch}. This learning paradigm disentangles exploration from exploitation, lending itself to the tasks in which exploration for collecting data is costly, time-consuming, or risky~\cite{isele2018safe, pmlr-v97-fujimoto19a}. By taking advantage of pre-collected datasets, offline RL can mitigate the technical concerns associated with online data collection, and has potential benefits for a number of real environments, such as human-robot collaboration~\cite{breazeal2008learning}. 

Offline reinforcement learning outsources the collection of data to a domain expert and subsequently searches for a batch-constrained optimal policy. However, this task is challenging, as offline RL methods suffer from the \emph{extrapolation error}~\cite{pmlr-v97-fujimoto19a, kumar2019stabilizing}. This pathology occurs when offline deep RL methods are trained under one distribution but evaluated on a different one. More specifically, value functions implemented by a function approximator have a tendency to predict unrealistic values for unseen state-action pairs for standard off-policy deep RL algorithms (\eg DQN and DDPG). This raises the need for approaches that restrict the action
space, forcing the agent to learn a behavior that is close to on-policy with respect to a subset of the given source data~\cite{pmlr-v97-fujimoto19a}.

For offline RL methods that are designed to mitigate the extrapolation error~\cite{pmlr-v97-fujimoto19a,kumar2019stabilizing,kumar2020conservative}, there remains the challenge that external data (\eg purchased from data markets) may not be well aligned with the intended target domain. Therefore, the learned policy induces a different visited state-action distribution that results in a degradation in the performance of the offline RL agent.

In recent years there have been a number of efforts within the paradigm of supervised learning for overcoming the \emph{source-target domain mismatch problem} via valuating data, including \emph{data Shapley}~\cite{ghorbani2019data} and \emph{data valuation using reinforcement learning} (DVRL)~\cite{yoon2020data}. Such methods have shown promising results on several application scenarios such as data-value quantification, corrupted sample discovery, robust learning with noisy labels, and domain adaptation~\cite{yoon2020data}. This raises the question: data valuation improve the transferability and robustness of an offline RL agent trained on externally acquired data?

To investigate the above question, we propose a data valuation approach that selects a subset of samples in the source dataset that are relevant to the target task. Our main contributions can be summarized as follows:
\begin{itemize}
    \item Inspired by DVRL~\cite{yoon2020data}, we propose Data Valuation for Offline Reinforcement Learning (DVORL) that for a given offline RL method, a fixed source dataset, and a small target dataset, identifies those samples of the source buffer that are relevant to the target task. 
    \item We contribute a benchmark on two Gym MuJoCo domains (Hopper and Walker2d) for which parameterizations (friction and mass of torso) for the target domain are different from those of the source domain.
    \item We show that the state-of-the-art offline RL methods fail to generalize to different target domain configurations.
    \item We show how our data valuation approach can improve the generalizability of the RL agent to the target domain by outperforming the existing state-of-the-art methods on all considered target domain configurations.
\end{itemize}

The rest of this paper is organized as follows. \autoref{sec:background} gives the background on (offline) RL. In \autoref{sec:problem_desc}, we formally define the source-target domain mismatch problem for offline RL, and provide a motivating example in \autoref{sec:motivating_example}. \autoref{sec:related_work} gives an overview of related work. In \autoref{sec:proposed_method}, we introduce our DVORL framework. \autoref{sec:experiments} describes our experiment setup. We discuss our results in \autoref{sec:results}, and in \autoref{sec:discussion}, we discuss our main findings. Finally in \autoref{sec:conclusion} we conclude with suggestions for future work.

\section{Background}
\label{sec:background}
\subsection{Reinforcement Learning}
The RL problem is typically modeled by a Markov decision process (MDP), formulated as a tuple $(\mathcal{X}, \mathcal{U}, p, r, \gamma)$, with a state space $\mathcal{X}$, an action space $\mathcal{U}$, and transition dynamics $p\left(x^{\prime} \mid x, u\right)$. 
At each discrete time step the agent performs an action $u \in \mathcal{U}$ in a state $x \in \mathcal{X}$, and transitions to a new state $x^{\prime} \in \mathcal{X}$ based on the transition dynamics $p\left(x^{\prime} \mid x, u\right)$, and receives a reward $r\left(x, u, x^{\prime}\right)$. The goal of the agent is to maximize the expectation of the sum of discounted rewards, also known as the return $R_{t}=\sum_{i=t+1}^{\infty} \gamma^{i} r\left(x_{i}, u_{i}, x_{i+1}\right)$, which weighs future rewards with respect to the discount factor $\gamma \in[0,1)$, determining the effective horizon.
The agent makes decisions via a policy $\pi: \mathcal{X} \rightarrow \mathcal{P}(\mathcal{U})$, which maps a given state $x$ to a probability distribution over the action space $\mathcal{U}$. For a given policy $\pi$, the value function is defined as the expected return of an agent, starting from state $x$, performing action $u$, and following the policy $Q^{\pi}(s, a)=\mathbb{E}_{\pi}\left[R_{t} \mid x, u\right]$. The state-action value function can be computed through the Bellman equation of the Q function:
\begin{equation}
Q^{\pi}(x, u)=\mathbb{E}_{s^{\prime} \sim p}\left[r\left(x, u, x^{\prime}\right)+\gamma \mathbb{E}_{u^{\prime} \sim \pi} Q^{\pi}(x^{\prime}, u^{\prime})\right].
\end{equation}
Given $Q^{\pi}$, the optimal policy $\pi^{*}=\operatorname{max}_{u} Q^{*}(x, u)$, can be obtained by greedy selection over the optimal value function $Q^{*}(x, u)=\max_{\pi} Q^{\pi}(x, u)$. 
For environments confronting agents with the curse of dimensionality the value can be estimated with a
differentiable function approximator $Q_\theta(x, u)$, with parameters $\theta$. 

In this work, we focus on continuous control problems, where, given a parameterized policy $\pi_{\vartheta}$
our objective is to find an optimal policy $\pi^*_{\vartheta}$, 
with respect to the parameters $\vartheta$, 
which maximizes the expected return 
from the start distribution
$J(\vartheta) = \E_{x_i \sim p_\pi, u_i \sim \pi} [R_0]$~\cite{lillicrap2015continuous}. 
The policy parameters $\vartheta$ 
can be updated by taking the gradient of the expected return $\nabla_{\vartheta}J(\vartheta)$. 
A popular approach to optimizing the policy is to use 
actor-critic methods, where the actor (policy) can be updated
through the deterministic policy gradient algorithm~\cite{silver2014deterministic}:
$\nabla_{\vartheta}J(\vartheta) = \E_{x \sim p_\pi}[\nabla_u Q_{\theta}^\pi(x, u) \rvert_{u, \pi(x)}\nabla_{\vartheta}\pi_{\vartheta}(x)]$, in which the value function $Q_{\theta}^\pi(x, u)$ is the critic.

\subsection{Offline Reinforcement Learning}
Standard off-policy deep RL algorithms such as deep Q-learning (DQN)~\cite{mnih2015human} and deep deterministic policy gradient (DDPG)~\cite{lillicrap2015continuous} are applicable in batch RL as they are based on more fundamental batch RL algorithms~\cite{fujimoto2019benchmarking}. However, they suffer from a phenomenon, known as \textit{extrapolation error}, which occurs when there is a mismatch between the given fixed batch of data and true state-action visitation of the current policy~\cite{pmlr-v97-fujimoto19a}. This is problematic as incorrect values of state-action pairs, which are not contained in the batch of data, are propagated through temporal difference updates of most off-policy algorithms~\cite{sutton1988learning}, resulting in poor performance of the model~\cite{thrun1993issues}. Below we give an overview of approaches designed
to address the extrapolation error, which will serve as our baselines.

\textbf{BCQ.} Batch-Constrained deep Q-learning~\cite{pmlr-v97-fujimoto19a} is an offline RL method for continuous control that restricts the action space, thereby eliminating actions that are unlikely to be selected by the behavioral policy $\pi_{b}$ and therefore rarely observed in the batch. Given a dataset of $N$ transitions $\mathcal{D}=$ $\left\{x_{t}, u_{t}, r_{t+1}, x_{t+1}\right\}_{t=0}^{N}$, collected by a behavior policy $\pi_{b}$, BCQ consists of four parameterized networks: 
a generative model $G_{\omega}: \mathcal{X} \rightarrow \mathcal{U}$, parameterized with $\omega$, a perturbation model $\xi_\phi(x, u, \Phi)$, parameterized with $\phi$, and two Q-networks $Q_{\vartheta_1}(x, u)$, $Q_{\vartheta_2}(x, u) $, parameterized with $\vartheta_1$ and $\vartheta_2$, respectively. 
The generative model $G_{\omega}$ selects the most likely action given the state with respect to the data in the batch. 
Since modeling the distribution of data in the high dimensional continuous control environments is not straightforward, a variational autoencoder (VAE) is used to approximate it. 
The policy is defined by sampling $n$ actions from $G_{\omega}(x)$ and selecting the highest valued action according to a Q-network as it is easier to sample from $\pi_{b}(u \mid x)$ than modeling $\pi_{b}(u \mid x)$ in a continuous action space. 
The perturbation model $\xi_{\phi}(x, u, \Phi)$, parameterized with $\phi$, models the distribution of data in the batch and is a residual added to the sampled actions in the range $[-\Phi, \Phi]$. 
This model is trained with the DDPG~\cite{silver2014deterministic} and can be thought of as a behavioral cloning model.
Since the perturbation model together with the sampling can be considered as a hierarchical policy, BCQ can also be considered an actor-critic method.

\textbf{CQL.} To prevent the training policy from overestimating the Q-values, Conservative Q-Learning (CQL)~\cite{kumar2020conservative} utilizes a penalized empircal RL objective. More precisely, CQL optimizes the value function not only to minimize the temporal difference error based on the interactions seen on the dataset but also minimizes the value of actions that the currently trained policy takes, while at the same time maximizing the value of actions taken by the behavioral policy during data generation. This results in a conservative $\mathrm{Q}$-function.

\textbf{TD3+BC.} Twin Delayed Deep Deterministic (TD3) policy gradient with Behavior Cloning (BC) is a model-free algorithm that does not explicitly learn a model of the behavioral policy, while trains a policy to mimic the behavior policy from the data~\cite{fujimoto2021minimalist}. It directly penalizes Euclidean distance to the actions that were recorded in the dataset.

\subsection{Data Valuation}
The training samples that machine learning (ML) models are trained with are not all equally valuable~\cite{lapedriza2013all}. 
A sample can be considered low-quality due to noisy input values, a noisy label, or the source-target domain mismatch problem. 
Removing low-quality samples has been shown to increase the performance of ML models~\cite{ghorbani2019data, yoon2020data}. The task of quantifying the quality of individual datum to the overall performance is referred to as \emph{data valuation}.
In supervised learning, data valuation is formally defined as follows. 
Given a source (training) dataset $\mathcal{D}_s=$ $\left\{\left(\mathbf{x}_{i}, y_{i}\right)\right\}_{i=1}^{N}$ and a target (test) dataset $\mathcal{D_{T}}=\left\{\left(\mathbf{x}_{j}^{\mathcal{T}}, y_{j}^{\mathcal{T}}\right)\right\}_{j=1}^{M}$ where $\mathbf{x} \in \mathcal{X}$ is a $d$-dimensional feature vector, and $y \in \mathcal{Y}$ is a corresponding label, the goal is to find a subset $\mathcal{D}^{*} = \left\{\left(\mathbf{x}_{k}, y_{k}\right) \mid \left(\mathbf{x}_{k}, y_{k}\right) \in \mathcal{D_S} \right\}_{k=1}^{K}$ of the source dataset $\mathcal{D_S}$ 
that maximizes the performance of the trained model on the target dataset $\mathcal{D_{T}}$~\cite{ghorbani2019data, durga2021training}.

\section{Problem Description} 
\label{sec:problem_desc}

In this section, we formally define the data valuation problem for offline RL.

We assume the availability of a source dataset \scalebox{1.}{$\mathcal{D_S}=$ $\{(x_{i}^{\mathcal{S}}, u^{\mathcal{S}}_{i}, {x^{\prime}_{i}}^{\mathcal{S}}, {r}_{i}^{\mathcal{S}}, {e}_{i}^{\mathcal{S}})\}_{i=1}^{N} \sim \mathcal{P}_{\mathcal{S}}$} and a
target dataset \scalebox{1.}{$\mathcal{D_T}=$ $\{(x_{i}^{\mathcal{T}}, u^{\mathcal{T}}_{i}, {x^{\prime}_{i}}^{\mathcal{T}}, {r}_{i}^{\mathcal{T}}, {e}_{i}^{\mathcal{T}})\}_{i=1}^{M} \sim \mathcal{P}_{\mathcal{T}}$}, 
where $x \in \mathbb{R}^{m}$ is a state; $u \in \mathbb{R}^{n}$ is the action that the agent performs at the state $x$; $r \in \mathbb{R}$ is the reward that the agent gets by performing the action $u$ in the state $x$; $x^{\prime} \in \mathbb{R}^{m}$ is the state that the agent transitions to (i.e. next state); and $e \in \{0, 1\}$ indicates whether the $x^{\prime}$ is a terminal state. 
We assume that the target dataset $\mathcal{D_T}$ is much smaller than the the source dataset $\mathcal{D_S}$, therefore $N \gg M$.
Furthermore, the source distribution $\mathcal{P}_{\mathcal{S}}$ can be different from the target distribution $\mathcal{P}_{\mathcal{T}}$, confronting our learner with the source-target domain mismatch problem. 
As in supervised learning, our goal is to find a (sub)set $\mathcal{D}^* \subseteq \mathcal{D_S}$, and seek a batch-constrained policy $\pi$, 
that when trained on $\mathcal{D}^*$ can generalize to the target domain used to construct $\mathcal{D_T}$.
Therefore, we have a transfer learning problem.

To formally define transfer learning for offline RL, we draw on the formulation from Zhu et al.~\cite{zhu2020transfer}. 
Let $\mathbf{\Psi}_{\mathcal{S}}=$ $\left\{{\Psi}_{\mathcal{S}} \mid {\Psi}_{\mathcal{S}} \in {\mathbf\Psi}_{\mathcal{S}}\right\}$ be a set of source domains and ${\Psi}_{\mathcal{T}}$ be a target domain, where each domain corresponds to an MDP. Therefore, the MDPs in the source domain $\Psi_{\mathcal{S}}$ and target domain $\Psi_{\mathcal{T}}$ are defined as $(\mathcal{X}_{\mathcal{S}}, \mathcal{U}_{\mathcal{S}}, p_{\mathcal{S}}, r_{\mathcal{S}}, \gamma_{\mathcal{S}})$ and $(\mathcal{X}_{\mathcal{T}}, \mathcal{U}_{\mathcal{T}}, p_{\mathcal{T}}, r_{\mathcal{T}}, \gamma_{\mathcal{T}})$, respectively. 
We assume prior knowledge $\mathcal{D}_{\mathcal{S}}$ provided by the set of source domains $\mathbf{\Psi}_{\mathcal{S}}$ and accessible to the target domain ${\Psi}_{\mathcal{T}}$.

By leveraging the prior information $\mathcal{D_{S}}$ from the source domain $\mathbf{\Psi}_{\mathcal{S}}$ as well as information $\mathcal{D}_{\mathcal{T}}$ provided by $\Psi_{\mathcal{T}}$, transfer learning aims to learn an optimal policy $\pi^{*}$ for the target domain ${\Psi}_{\mathcal{T}}$, such that:
\begin{equation}
\pi^{*}=\underset{\pi}{\arg \max } \mathbb{E}_{x \sim \mathcal{X_{T}}, u \sim \pi}\left[Q_{\Psi_{\mathcal{T}}}^{\pi}(x, u)\right],
\end{equation}
where $\pi=\phi\left(\mathcal{D_{S}} \sim \Psi_{\mathcal{S}}, \mathcal{D}_{\mathcal{T}} \sim \Psi_{\mathcal{T}}\right): \mathcal{X}_{\mathcal{T}} \rightarrow \mathcal{U}_{\mathcal{T}}$ is a function mapping the states to actions for the target domain $\Psi_{\mathcal{T}}$, learned based on information from both $\mathcal{D}_{\mathcal{T}}$ and $\mathcal{D}_{\mathcal{S}}$.

The source and target domains can have distinct state spaces, but their action spaces have to be the same and their transition function and reward function have to be similar as they share internal dynamics. We focus on policy transfers where $\mathcal{X}_{\mathcal{S}} = \mathcal{X}_{\mathcal{T}}$, 
$\mathcal{U}_{\mathcal{S}}=\mathcal{U}_{\mathcal{T}}$, 
$r_{\mathcal{S}} = r_{\mathcal{T}}$,
but $p_{\mathcal{S}} \neq p_{\mathcal{T}}$.

The source and target domains can have distinct transition functions because a change in environment parameters (\eg mass of torso and friction) results in a different probability function $p$, which itself is conditioned on $(x, u, x^{\prime})$, where $x$, $u$, and $x^{\prime}$ denote state, action, and next state, respectively.

However, since the target dataset is very small, compared to the source dataset ($N \gg M$), state-action transition $(x, u, x^{\prime})$ is too restrictive. Thus, we also consider the case $(x)$ in which only the change in the distribution of the state space is taken into account.

\section{Motivating example} 
\label{sec:motivating_example}

\begin{figure*}
\center
\includegraphics[width=0.97\textwidth]{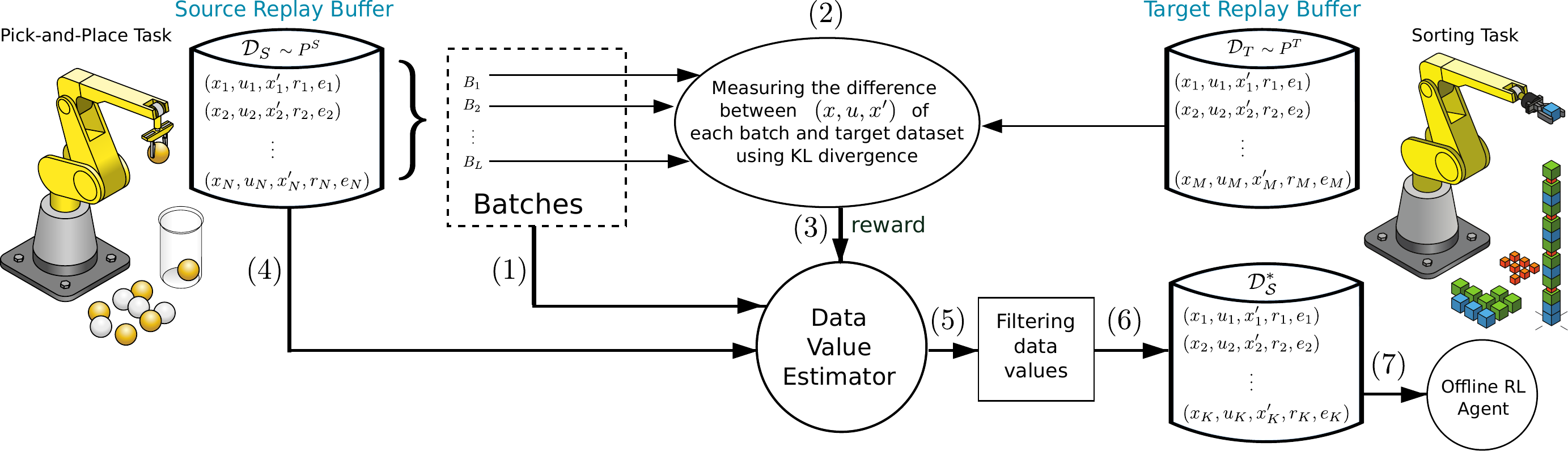}
\caption{Illustration of the DVORL framework. (1) A batch of source buffer samples is given as input to the data value estimator (DVE). (2) KL Divergence between the distribution of the state-action transitions of the given batch and the distribution of the state-action transitions of target buffer (whose transition items are collected by a domain expert on the target domain) is calculated and (3) used as the reward signal for updating the DVE. (4) After all the samples of the source buffer are used for training the DVE, (5) the fully trained DVE is used for outputting the data values of the source buffer (6) that are filtered out by removing those values being lower than the selection threshold. (7) This results in a subset of the source buffer that is relevant to the target domain and is used for training the given base offline RL algorithm.}

\label{diagaram}
\end{figure*}

For our motivating example, we consider two scenarios involving a cobot, depicted in \autoref{diagaram}. 
In the source domain the cobot is performing pick-and-place task while the target domain confronts the cobot with a sorting task. Clearly, a learning agent trained on the source task will perform poorly on the target task. However, our hypothesis is that data valuation can help us identify samples that are relevant for both tasks. For instance, both tasks have pick-up and place actions in common. Therefore, the goal is to find the relevant subset of the source dataset that allows the agent to learn a policy that generalizes to the target domain.

\section{Related Work}
\label{sec:related_work}

The RL literature contains numerous techniques for dealing with the source-target domain mismatch problem.
Noteworthy contributions here include:
EPOPT~\cite{rajeswaran2016epopt}, 
which is a combination of policy transfer through source domain ensemble and learning from limited demonstrations for the fast adaptation to the target domain; 
UP-OSI~\cite{yu2017preparing} trains robust agent policies using a large number of synthetic demonstrations from a simulator to deal with environments with unknown dynamics; 
CAD$^2$RL~\cite{sadeghi2016cad2rl} learns latent representations
from observations in the source domain that are generally applicable to the target domain; 
DARLA~\cite{higgins2017darla}, a zero-shot transfer learning method that learns disentangled representations which are robust against domain shifts; 
and SAFER~\cite{slack2022safer}, which accelerates policy learning on complex control tasks by considering safety constraints.

Meanwhile, the literature on off-policy RL includes principled experience replay memory sampling techniques. Prioritized Experience Replay (PER)~\cite{schaul2015prioritized} (e.g., \cite{hou2017novel, horgan2018distributed, kang2021deep}) attempts to sample transitions that contribute the most toward learning.
prioritized replay with weighted importance sampling
can improve BCQ.
However, the majority of the work to date on offline RL is focused on preventing the training policy from being too disjoint with the behavioral policy~\cite{pmlr-v97-fujimoto19a, kumar2020conservative, kidambi2020morel}. 
To increase the generalization capacity of offline RL methods, Kostrikov et al.~\cite{kostrikov2021offline} propose in-sample Q-learning (IQL), which approximates the policy improvement step by considering the state value function as a random variable with some randomness determined by the action, and then taking a state-conditional expectile of this random variable to estimate the value of the best actions in that state.

In contrast to the offline RL methods listed above, our work focuses on valuating the suitability of state transition tuples for a given target domain.
Here we draw inspiration from the literature on data valuation for supervised learning.
Ghorbani et al. in \cite{ghorbani2020distributional} propose the distributional Shapley, which is a framework in which the value of a point is defined in the context of underlying data distribution. The reformulation of the data Shapley value as a distributional quantity reduces the dependence on the specific draw of data as the valuation function does not depend on a fixed data set. 
Ghorbani and Zou propose the Neuron Shapley framework~\cite{ghorbani2020neuron} to quantify how individual neurons contribute to the prediction and performance of a deep neural network (DNN). However, the limitation of Shapely based methods is that it's computationally expensive or even impossible to quantify the contribution of each individual sample to the overall performance of the model, in particular, for complex models such as DNNs~\cite{yoon2020data}.

Wang et al. \cite{wang2019minimax} propose a minimax game-based transfer learning technique for selective transfer learning that consists of a selector, a discriminator, and a transfer learning module, playing a minimax game to find useful source data. 
Yoon et al.~\cite{yoon2020data} introduce a framework for data valuation in supervised learning tasks, making use of RL to determine how likely each training sample is used in the training of the predictor model. 
This method integrates data valuation into the training procedure of the predictor model, making the predictor and data value estimator able to improve each other’s performance. 

Despite the success of the existing works on data valuation, they are only applicable to the supervised learning tasks in which the availability of labels is assumed. Therefore, they are not directly applicable to the RL setting where there is no ground-truth for the transitions. 
This emphasizes a need for a data valuation method that is applicable to the RL setting.

\section{Proposed Method}
\label{sec:proposed_method}
DVORL consists of two learnable functions: (1) a data value estimator (DVE) model {$v_{\phi}$} and (2) an offline reinforcement learning model. 
Inspired by DVRL~\cite{yoon2020data}, we adapt the REINFORCE algorithm~\cite{williams1992simple} and use it as the DVE. 
We use a DNN for \added[id=GP]{the} DVE.
The goal is to find the parameters \added[id=GP]{$\phi^*$} of the DNN so that the network returns the optimal probability distribution over the sample space. 


The DVE model $v_{\phi}: \mathcal{X} \times \mathcal{U} \times \mathcal{X^{\prime}} \times  \mathcal{R} \times \mathcal{E} \rightarrow[0,1]$ is optimized to output data values corresponding to the relevance of training samples to the target task. We formulate the corresponding optimization problem as:
\begin{equation}
\resizebox{0.90\hsize}{!}{%
$\max _{\phi} J\left(\pi_{\phi}\right) =  \E_{\substack{(x^{\mathcal{S}}, u^{\mathcal{S}}, x^{'\mathcal{S}}) \sim P^{\mathcal{S}}\\(x^{\mathcal{T}}, u^{\mathcal{T}}, x^{'\mathcal{T}}) \sim P^{\mathcal{T}}}}\left[r_{\phi}((x^{\mathcal{S}}, u^{\mathcal{S}}, x^{'\mathcal{S}}), (x^{\mathcal{T}}, u^{\mathcal{T}}, x^{'\mathcal{T}})) \right]$,
}
\end{equation}
where 
\begin{equation}
r_{\phi} = \frac{1}{ \infdiv{\mathcal{P}_{(x^{\mathcal{S}}, u^{\mathcal{S}}, x^{'\mathcal{S}})}}{\mathcal{P}_{(x^{\mathcal{T}}, u^{\mathcal{T}}, x^{'\mathcal{T}})}}}
\end{equation}
corresponds to the reciprocal of the KL divergence between the batch of source dataset and target dataset. 
Therefore, the objective of the network is to assign high probabilities to samples whose reward function value $r_{\phi}((x^{\mathcal{S}}, u^{\mathcal{S}}, x^{'\mathcal{S}}), (x^{\mathcal{T}}, u^{\mathcal{T}}, x^{'\mathcal{T}}))$ is high.

\vspace{2mm}
\textbf{Training.} As shown in \autoref{alg:NUI-DDQN} (lines 4 to 10), all the samples of source buffer $\mathcal{D_S}$ are divided into batches and each batch $\mathcal{D_S}'$ is given as input to the DVE (with shared parameters across the batch). 
%
The KL divergence between the distribution of the state-action transition $(x, u, x^{'})$ of the given batch and that of the small target buffer $\mathcal{D_T}$ is calculated and used as the reward signal $r_{sig}$ for training the DVE. 
Let $w=v_{\phi}(x_{i}^{\mathcal{S}}, u^{\mathcal{S}}_{i}, {x^{\prime}_{i}}^{\mathcal{S}}, {r}_{i}^{\mathcal{S}}, {e}_{i}^{\mathcal{S}})$ denote the probability that the sample $i$ of the source buffer is used for training the offline reinforcement learning model.

Our adapted version of REINFORCE algorithm, has the following object function for the policy $\pi_{\phi}$:

\vspace{1mm}
\resizebox{0.97\hsize}{!}{%
\begin{math}
\begin{aligned}
J\left(\pi_{\phi}\right)&= \E_{\substack{\substack{(x^{\mathcal{S}}, u^{\mathcal{S}}, x^{'\mathcal{S}}) \sim P^{\mathcal{S}}\\(x^{\mathcal{T}}, u^{\mathcal{T}}, x^{'\mathcal{T}}) \sim P^{\mathcal{T}} \\ w \sim \pi_{\phi}(\mathcal{D_S}', \cdot)}}}\left[r_{\phi}((x^{\mathcal{S}}, u^{\mathcal{S}}, x^{'\mathcal{S}}), (x^{\mathcal{T}}, u^{\mathcal{T}}, x^{'\mathcal{T}}))\right] \\
&=\int P^{\mathcal{T}}\left((x^{\mathcal{S}}, u^{\mathcal{S}}, x^{'\mathcal{S}})\right) \sum_{w \in[0,1]^{N}} \pi_{\phi}(\mathcal{D_S}', w) \\
& \quad\cdot\left[r_{\phi}((x^{\mathcal{S}}, u^{\mathcal{S}}, x^{'\mathcal{S}}), (x^{\mathcal{T}}, u^{\mathcal{T}}, x^{'\mathcal{T}})) \right] d \left((x^{\mathcal{S}}, u^{\mathcal{S}}, x^{'\mathcal{S}})\right).
\end{aligned}
\end{math}
}

\vspace{1mm}
In the above equation, $\pi_{\phi}(\mathcal{D_S}', w)$ is the probability that the selection probability vector $w$ occurs. In this way, the policy directly uses the values output by the DVE. This is different from the DVRL~\cite{yoon2020data}, which uses a binary selection vector $\mathbf{s}=\left(s_{1}, \ldots, s_{B_{s}}\right)$ where $s_{B_{s}}$ denotes the batch size, $s_i \in \{0,1\}$, and $P\left(s_{i}=1\right)=w_{i}$. Thus, in our training, the DVE has no control over exploration and just provides weightings for the given samples and is tuned accordingly.

It should be noted that we use the whole input information of the source buffer (i.e., $\left(x, u, x^{\prime}, {r}, {e}\right)$) for calculating the data values (i.e., $\mathcal{D_S}'$ in policy $\pi_{\phi}(\mathcal{D_S}', \mathbf{w})$); however, we only use the information of the state-action transition ($x, u, x^{\prime}$) for calculating the reward signal, used for updating the DVE, that is consistent with our formulation of transfer learning where $p_{\mathcal{S}} \neq p_{\mathcal{T}}$ as the transition probability function is conditioned on the state-action transition ($x, u, x^{\prime}$).

We calculate the gradient of the above objective function in the following.

\vspace{1mm}
\resizebox{0.97\hsize}{!}{%
\begin{math}
\begin{aligned}
    \nabla_{\phi} J\left(\pi_{\phi}\right) &=\int P^{T}\left((x^{\mathcal{S}}, u^{\mathcal{S}}, x^{'\mathcal{S}})\right) \sum_{w \in[0,1]^{N}} \nabla_{\phi} \pi_{\phi}(\mathcal{D_S}', w) \\
    &\quad \cdot\left[r_{\phi}((x^{\mathcal{S}}, u^{\mathcal{S}}, x^{'\mathcal{S}}), (x^{\mathcal{T}}, u^{\mathcal{T}}, x^{'\mathcal{T}}))) \right] d \left((x^{\mathcal{S}}, u^{\mathcal{S}}, x^{'\mathcal{S}})\right)\\
    &=\int P^{T}\left((x^{\mathcal{S}}, u^{\mathcal{S}}, x^{'\mathcal{S}})\right)\left[\sum_{x \in[0,1]^{N}} \frac{\nabla_{\phi} \pi_{\phi}(\mathcal{D_S}', w)}{\pi_{\phi}(\mathcal{D_S}', w)} \right.\\
    &\quad\left.\cdot\;\pi_{\phi}(\mathcal{D_S}', w) \left[r_{\phi}((x^{\mathcal{S}}, u^{\mathcal{S}}, x^{'\mathcal{S}}), (x^{\mathcal{T}}, u^{\mathcal{T}}, x^{'\mathcal{T}})) \right]\vphantom{\sum_{x \in[0,1]^{N}}}\right] d \left((x^{\mathcal{S}}, u^{\mathcal{S}}, x^{'\mathcal{S}})\right)\\
    &=\int P^{T}\left((x^{\mathcal{S}}, u^{\mathcal{S}}, x^{'\mathcal{S}})\right)\left[\sum_{w \in[0,1]^{N}} \nabla_{\phi} \log \left(\pi_{\phi}(\mathcal{D_S}', w)\right) \right. \\
    &\quad\left. \cdot\;\pi_{\phi}(\mathcal{D_S}', w) \left[r_{\phi}((x^{\mathcal{S}}, u^{\mathcal{S}}, x^{'\mathcal{S}}), (x^{\mathcal{T}}, u^{\mathcal{T}}, x^{'\mathcal{T}})) \right]\vphantom{\sum_{w \in[0,1]^{N}}}\right]  d \left((x^{\mathcal{S}}, u^{\mathcal{S}}, x^{'\mathcal{S}})\right)\\
    &=\underset{\substack{(x^{\mathcal{S}}, u^{\mathcal{S}}, x^{'\mathcal{S}}) \sim P^{S}\\(x^{\mathcal{T}}, u^{\mathcal{T}}, x^{'\mathcal{T}}) \sim P^{T} \\ w \sim \pi_{\phi}(\mathcal{D}_S, \cdot)}}{\mathbb{E}}\left[r_{\phi}((x^{\mathcal{S}}, u^{\mathcal{S}}, x^{'\mathcal{S}}), (x^{\mathcal{T}}, u^{\mathcal{T}}, x^{'\mathcal{T}})) \right] \\
    &\quad\cdot \nabla_{\phi} \log \left(\pi_{\phi}(\mathcal{D_S}', w)\right).
\end{aligned}
\end{math}
}

\vspace{2mm}
To enhance the stability of the DVE, we use the moving average $r_{rolling}$ of the previous signal rewards with the window size $\omega$ as the baseline. The baseline reduces the variance of the gradient estimates~\cite{greensmith2004variance}. 

\vspace{2mm}
\textbf{Inference.} 
As shown in \autoref{alg:NUI-DDQN} (lines 11 to 18), after all the samples of the source buffer are used for training the DVE, the fully-trained DVE is used for outputting the data values of the original source buffer. The samples whose corresponding data values are lower than the selection threshold $\epsilon$ are filtered out and the remaining subset of samples form the new source buffer $\mathcal{D_S^*}$ that is relevant to the target domain:
\begin{equation}
\resizebox{0.90\hsize}{!}{%
$\mathcal{D_S^*}=\left\{\left(x_{i}^{\mathcal{S}}, u^{\mathcal{S}}_{i}, {x^{\prime}_{i}}^{\mathcal{S}}, {r}_{i}^{\mathcal{S}}, {e}_{i}^{\mathcal{S}}\right) \in \mathcal{D_S} \;|\; i=1, \ldots, N; w_i \geq \epsilon \right\}$.
}
\end{equation}

Finally, the buffer $\mathcal{D_S^*}$ is given to an offline RL model for training.

\begin{algorithm}[ht]
\caption{Data Valuation for Offline RL}
\label{alg:NUI-DDQN}
  \begin{algorithmic}[1]
    \State \textbf{Input:} Fixed source and target buffers $\mathcal{D_S}$ and $\mathcal{D_T}$; mini-batch size $B_s>$ 0; learning rate $\alpha$; moving average window size $\omega$; selection threshold $\epsilon$.
    \State \textbf{Output:} A subset of source buffer $\mathcal{D_S^*}$ which is relevant to target domain.
    \State \textbf{Initialize:} $\mathcal{D_S^*} \gets \varnothing$; DVE parameters $\phi$; $r_{\phi} \gets 0$.    
    \For {batch $B_j$ in $\mathcal{D_S}$}
        \State $w_j = v_{\phi}(B_j)  =  v_{\phi}(x_j, u_j, x^{\prime}_j, {r_j}, {e_j})$ 
        \State $r_{\phi} \gets \frac{1}{ \infdiv{\mathcal{P}_{(x^{\mathcal{S}}, u^{\mathcal{S}}, s^{'\mathcal{S}})}}{\mathcal{P}_{(x^{\mathcal{T}}, u^{\mathcal{T}}, s^{'\mathcal{T}})}}}$ 
    \State $r_{sig} = r_{\phi} - r_{rolling}$ 
    \State $\phi \gets \phi - \alpha \left[r_{sig} \right] \nabla_{\phi} \log \pi_{\phi}\left(B_j, w_j\right)$
    \State $r_{rolling} \gets \frac{\omega-1}{\omega} r_{rolling}+\frac{1}{\omega} r_{\phi}$
    \EndFor  
    \For {batch $B_k$ in $\mathcal{D_S}$} 
    \State $w_k = v_{\phi}(B_k) = v_{\phi}(x_k, u_k, x^{\prime}_k, {r_k}, {e_k})$ 
        \For {sample $i$ in $B_k$}
            \If{$w_i \geq \epsilon$} 
                \State $\mathcal{D_S^*} = \mathcal{D_S^*} \cup (x_i, u_i, x^{\prime}_i, {r_i}, {e_i})$
            \EndIf
        \EndFor
    \EndFor \\
    \Return $\mathcal{D_S^*}$
\end{algorithmic}
\end{algorithm}

\section{Experiments}
\label{sec:experiments}
\subsection{Baselines}
In this work, we use the following offline RL methods discussed in \autoref{sec:background} as baselines: (Vanilla) BCQ, CQL and TD3\replaced[id=GP]{+}{\_}BC. We also evaluated the performance of BEAR~\cite{kumar2019stabilizing} on our considered domains. However, as found by~\cite{kumar2020conservative}, we found CQL and TD3\replaced[id=GP]{+}{\_}BC outperformed BEAR, and therefore focused our evaluation on the above three methods and our DVORL.

We use the (Vanilla) BCQ as the base model in our DVORL, and we refer to it as Data Valuation based BCQ (DVBCQ). The reason for using the Vanilla BCQ is that it is the most commonly used offline RL algorithm, and we also intend to show how the selection of relevant transitions can help a base model, underperforming other methods in most cases, outperform the state-of-the-art methods in terms of transferability of learned policy to different target configurations. We consider two versions of DVBCQ: i)~DVBCQ~$(x)$ using information of states, and ii)~DVBCQ $(x, u, x^{'})$ using information of state-action transition, for calculating the similarity between source and target buffers.

\subsection{Gathering samples for the replay buffer}


The DVORL agent learns from a dataset collected by a domain expert. In our experiments, for each domain and domain parametrization, we trained a DDPG agent for one million iterations and used the fully-trained agent for generating the buffers with the size one million and ten thousand for the source and target, respectively.


\subsection{Domains}

In this work, we use the following two MuJoCo domains:
\begin{itemize}
    \item \textbf{Hopper-v3:} The Hopper is a simulated monopod robot with 4 body links including the torso, upper leg, lower leg,
and foot, together with 3 actuated joints. This domain has an 11-dimensional state space
including joint angles and joint velocities and a 3-dimensional action space corresponding to torques at the joints. The goal is to make the hopper hop forward as fast as possible.
    \item \textbf{Walker2d-v3:} The Walker is a simulated bipedal robot consisting of 7 body links including to two legs and a torso, along with 6 actuated joints. This domain has a 17-dimensional state space including joint angles and joint velocities and a 4-dimensional action space corresponding to joint torques. The goal is to make the robot walk forward as fast as possible.
\end{itemize}

\begin{table}[htbp!]
\caption{Domain configurations.}
\label{tab:confs}
  \centering
\begin{adjustbox}{width=0.9\columnwidth}
\begin{tabular}{l c c c}
    \toprule
     {Domain} & {Source Config} & {Target Config}  & {Name}\\ \midrule
    {Hopper-v3} & \shortstack{F: 2.0\\\;\,T: 0.05} & \shortstack{F: 2.0\\\;\,T: 0.05} & {Hopper-Source} \\
    \cmidrule(l){3-4}
    {} & {} & \shortstack{F: 2.5\\\;\;\;\,T: 0.075} & {Hopper-Target1} \\
    \cmidrule(l){3-4}
    {} & {} & \shortstack{F: 3.0\\\;\;\;\,T: 0.075} & {Hopper-Target2} \\
    \midrule
    {Walker2d-v3} & \shortstack{F: 0.9\\\;\,T: 0.05} & \shortstack{F: 0.9\\\;\,T: 0.05} & {Walker2d-Source} \\
    \cmidrule(l){3-4}
    {} & {} & \shortstack{F: 1.125\\T: 0.075} & {Walker2d-Target1} \\
    \cmidrule(l){3-4}
    {} & {} & \shortstack{F: 1.35\\\;\,T: 0.075} & {Walker2d-Target2} \\
    \bottomrule
\end{tabular}
\end{adjustbox}
\end{table}

\subsection{Source and target domain settings}
\label{domain-settings}
For our experiments, we shall distinguish between \emph{source} and \emph{target} domains. The source domain is the one within which the samples are gathered by a fully-trained DDPG. The target domain is the domain within which the DVORL agent is to be deployed. To study the extent to which DVBCQ can cope with modified domain configurations, we consider two scenarios with respect to the source and target domains:

\begin{enumerate}
    \item \textbf{Identical Source and Target Domains:} Domain configuration for gathering samples and training DVORL agent remain the same. This is the simplest setting where the DDPG agent gathers samples in an environment with a domain parameterization identical to the domain within which the DVORL agent will be deployed. For this setting, we consider two datasets ``Hopper-Source" and ``Walker-Source". 
    \item \textbf{Transfer Learning:} Samples are gathered from a source domain with a parameterization that differs from the target domain. More precisely, the target domain will have different \emph{mass of torso} and  \emph{friction} coefficients compared with the source domain. For this setting, we consider four datasets ``Hopper-Target1", ``Hopper-Target2", ``Walker2d-Target1", and ``Walker2d-Target2". 
\end{enumerate}
All the considered source and target domain configurations are presented in \autoref{tab:confs}.



\subsection{Parameter Tuning}
For all the competitors, we used the default parameters values reported in the corresponding papers. Hyperparameters of DVORL are selected by grid search. Since we used the BCQ in our DVORL method, we report the used parameter values of Data Valuation based BCQ (DVBCQ), listed in \autoref{tab:parametes}. The parameters values of the baseline (Vanilla) BCQ are the same as those of the base agent in our DVBCQ.

\begin{table}[htbp!]
\caption{Parameters values of our DVBCQ model.}
\label{tab:parametes}
  \centering
  \begin{adjustbox}{width=\columnwidth}
\begin{tabular}{l c l}
    \toprule
    {Parameter} & {Value} & {Description}  \\ \midrule
    {dve\_batch\_size}  & {200} & {batch size for DVE}  \\
    {dve\_hidden\_layers}  & {[128, 128]} & {Number of nodes in hidden layers} \\
    {moving\_average\_window\_size}  & {20} & {Window size for the moving average} \\
    {selection\_threshold}  &  {0.1}  & {selection threshold} \\
    {mini\_batch\_size}  &  {100}  & {Batch size for offline RL model} \\
    {discount}  &  {0.9}  & {Discount factor} \\
    {tau}  &  {0.005}  & {Target network update rate of BCQ} \\
    {lambda}  &  {0.75}  & {Weighting for clipped double Q-learning} \\
    {phi}  &  {0.05}  & {Max perturbation parameter for BCQ} \\
     \bottomrule
\end{tabular}
\end{adjustbox}
\end{table}

\subsection{Implementation}
Our implementation of the DVORL builds
on OpenAI gym’s~\cite{brockman2016openai} control environments with the MuJoCo~\cite{todorov2012mujoco} physics simulator.

\section{Results}
\label{sec:results}

\subsection{Evaluation of offline RL methods on source and target domains}
\begin{figure*}[!ht]
  \centering
    \resizebox{\textwidth}{!}{
  \includegraphics[width=0.9\textwidth]{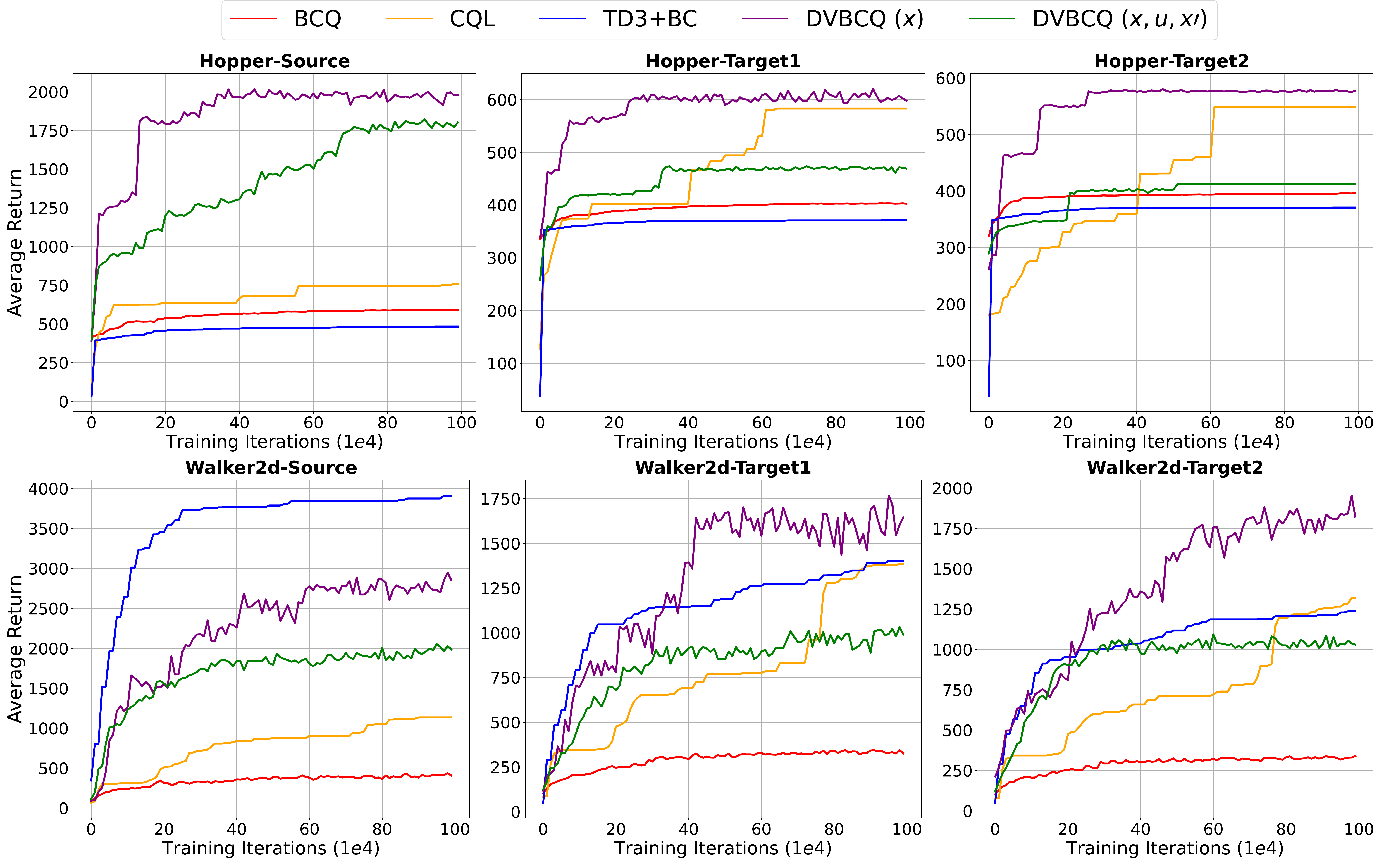}
  }
  \caption{Performance of BCQ, CQL, TD3\replaced[id=GP]{+}{\_}BC, and DVBCQ on the source domain and two different target domain configurations (described in  \autoref{domain-settings}), where the models are trained for 1M iterations, and the results are averaged over ten runs on target domain environments with randomly-chosen seeds. For DVBCQ and other baselines, we report the performance achieved by the best policy with respect to checkpoints saved throughout the run.}
\label{fig:models_perf}
\end{figure*}

\autoref{fig:models_perf} shows the performance of BCQ, CQL, TD3+BC, DVBCQ $(x)$ and DVBCQ $(x, u, x^{'})$ on the source domain and two different target domain configurations described in  \autoref{domain-settings}. The models are trained for 1M iterations, and the results are averaged over ten runs on target domain environments with randomly-chosen seeds. 

\noindent{\textit{Identical source-domain:}} For DVBCQ and other baselines, we report the average return achieved by the best policy with respect to checkpoints saved throughout the run. For the identical source-domain setting, both DVBCQ $(x)$ and  DVBCQ $(x, u, x^{'})$ significantly outperform all baselines on Hopper environment, and their performance is superior to BCQ and CQL, while underperforming TD3+BC on Walker2d environment. 

\noindent{\textit{Transfer learning:}}  For transfer learning setting, DVBCQ $(x)$ outperforms both target domains whose configurations (mass of torso and friction) differ from those of the source domain, on both Hopper and Walker2d environments. However, DVBCQ $(x, u, x^{'})$ underperforms CQL on Hopper environment and TD3+BC on Walker environment but it has competitive performance compared with other baselines.

It should be noted that the there is a significant difference between the performance of DVBCQ and its base model BCQ.

\subsection{Removing high/low value transitions from source dataset}
\begin{figure}[!h]
  \centering
    \resizebox{.45\textwidth}{!}{
  \includegraphics[width=\textwidth]{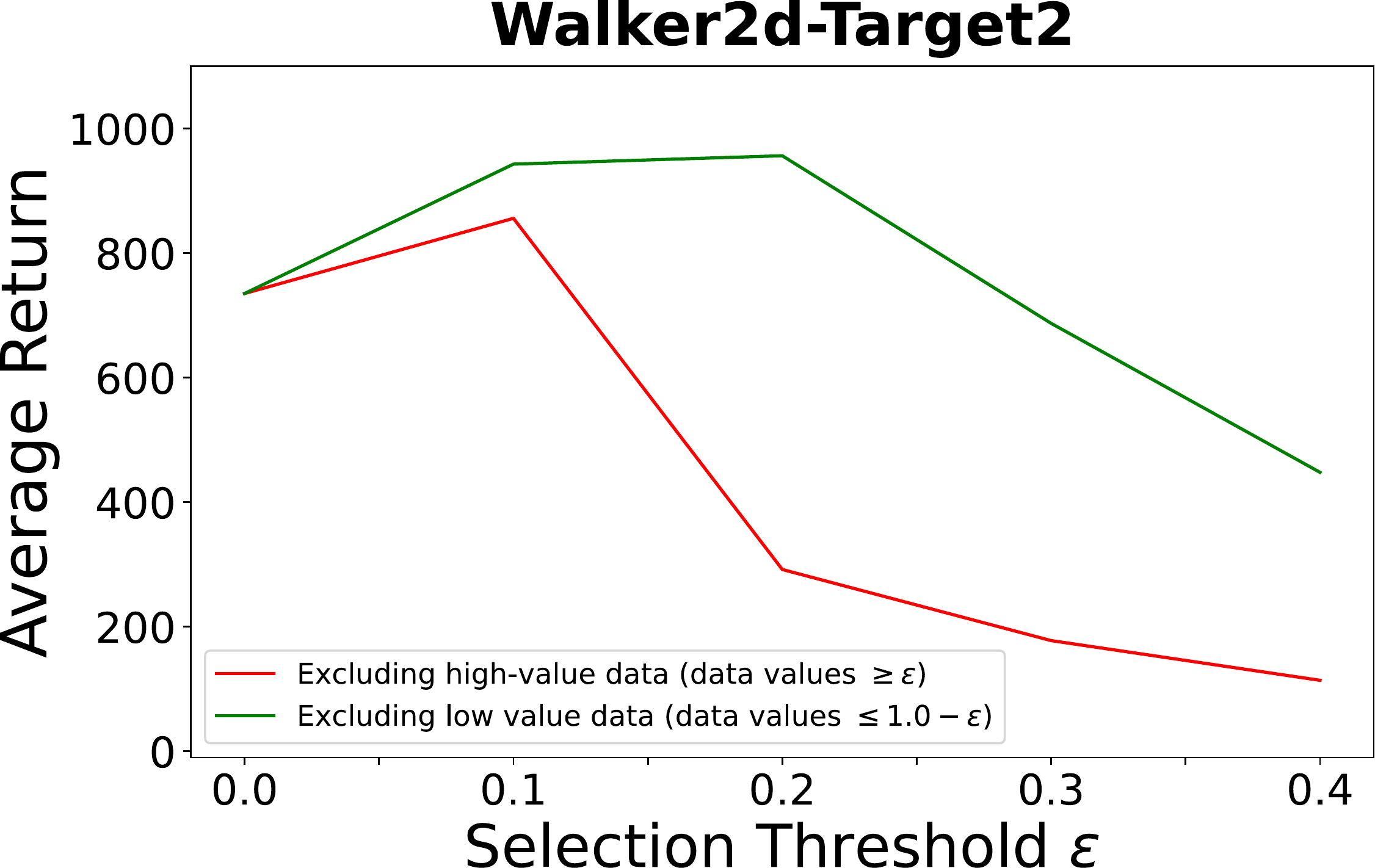}}
  \caption{Performance of BCQ models trained on the source
dataset with different selection thresholds and evaluated on a different target domain configuration (Walker2d-Target2), where all the models are trained for 200.000 iterations, and a fixed seed is used for the evaluation environment. Excluding high-value samples (red line) aggravates the performance of the offline RL methods. However, excluding low-value samples (green line) does not deteriorate the performance as much as that of the high-value samples.}
  \label{fig:exclude_threshold}
\end{figure}

\autoref{fig:exclude_threshold} shows the performance of BCQ models trained on the source dataset with different selection thresholds and evaluated on a different target domain configuration (Walker2d-Target2), where all the models are trained for 200K iterations, and a fixed seed is used for the evaluation environment. We consider five thresholds (0.0, 0.1, ..., 0.4) for excluding the high/low-value data samples of the source dataset. In addition, we report the average return of the best policy (with respect to checkpoints saved throughout the run) learned for each point.

As shown in \autoref{fig:exclude_threshold},  removing low-value samples from the source dataset can help the RL agent learn only those transitions relevant to the target domain configuration and therefore achieve better performance on the target domain (green line). On the other hand, removing high-value samples from the source dataset significantly deteriorates the RL agent's performance (red line). 

The findings in \autoref{fig:exclude_threshold} support the opening hypothesis that excluding high-value samples worsens the performance of the offline RL methods.

\section{Discussion \& Future Work}
\label{sec:discussion}
Our results suggest that DVORL can improve the offline reinforcement learning methods on both identical source-target and transfer learning settings. In addition, our method helps the offline RL methods achieve significantly higher performance with fewer iterations, making them more efficient. Furthermore, our method can identify the relevant samples of the source domain to different target domain configurations. This is of high importance and has many use cases, such as learning from an externally acquired dataset and safe RL.

It should be noted that our goal is not to show that our proposed method outperforms all the state-of-the-art offline RL methods on both source and target domains, but to show that the data valuation for the offline reinforcement learning (DVORL) framework can improve the performance of the baseline algorithms.

For future work, we aim to examine whether the size of target buffer plays a role in the performance of DVORL. We intend conduct some experiments on real-world domains and compare our results to other data valuation methods like Data Shapely. Moreover, we plan to improve our reward function by taking into account dynamics of the model. 

We also aim to investigate the extent to which DVORL can identify the safe transitions within a safe reinforcement learning setting. We also plan to apply the idea of transition valuation to the safe multi-agent reinforcement learning~\cite{elsayed2021safe}, where different data value estimators are optimized for the corresponding agent with respect to the tasks that they need to perform. In addition, we aim to incorporate a mechanism for auto-tuning the selection threshold into the training as the optimal value for this parameter may vary from one domain configuration to another one.
\section{Conclusion}
\label{sec:conclusion}
In this work, we proposed a data valuation framework that selects a subset of samples in the source dataset that are relevant to the target task. The data values are estimated using a deep neural network, trained using reinforcement learning with a reward that corresponds to the similarity between the distribution of the state-action transition of the given data and the target dataset. 
We show that DVORL outperforms baselines on different target domain configurations and has a competitive performance on the source domain in which the reinforcement learning agent is trained. We find that our method can identify relevant and high-quality transitions and improve the performance and transferability of policy learned by offline RL algorithms. Moreover, we contributed a benchmark on two Gym MuJoCo domains (Hopper and Walker2d) for which domain configurations (friction and mass of torso) for the target domain differ from those of the source domain.







\section*{ACKNOWLEDGMENT}
The authors gratefully acknowledge, that the proposed research is a result of the
research project “IIP-Ecosphere”, granted by the German Federal Ministry for
Economics and Climate Action (BMWK) via funding code 01MK20006A.


\bibliographystyle{IEEEtran} 

\bibliography{references}

\end{document}